# Enhancing Navigation Benchmarking and Perception Data Generation for Row-based Crops in Simulation

Martini M*., Eirale A., Tuberga B., Ambrosio M., Ostuni A., Messina F., Mazzara L., Chiaberge M.
*Department of Electronics and Telecommunication (DET), Politecnico di Torino, Corso Duca degli Abruzzi 24, Torino (TO), 10129*
* mauro.martini@polito.it

**Abstract**

Service robotics is recently enhancing precision agriculture enabling many automated processes based on efficient autonomous navigation solutions. However, data generation and infield validation campaigns hinder the progress of large-scale autonomous platforms. Simulated environments and deep visual perception are spreading as successful tools to speed up the development of robust navigation with low-cost RGB-D cameras. In this context, the contribution of this work is twofold: a synthetic dataset to train deep semantic segmentation networks together with a collection of virtual scenarios for a fast evaluation of navigation algorithms. Moreover, an automatic parametric approach is developed to explore different field geometries and features. The simulation framework and the dataset have been evaluated by training a deep segmentation network on different crops and benchmarking the resulting navigation.

## Introduction

In the last two decades, scientific research on mobile robots operating in agricultural environments has greatly expanded. In this field, four essential requirements have been identified: increasing productivity, allocating resources reasonably, adapting to climate change, and avoiding food waste (Zhai et al., 2020). Assistive robotics platforms may play a significant role in many agricultural tasks, such as harvesting (Bac et al., 2014), spraying (Deshmukh et al., 2021), vegetative assessment, yield estimation (Feng et al., 2020; Zhang et al., 2020), since they reduce human labor and enhance operational safety. One fundamental aspect necessary to operate all these tasks is a robust and reliable navigation system. However, autonomous navigation in agricultural environments introduces several peculiar challenges due to weather or lighting conditions, irregular terrain, and plant vegetation. Localization in crop field is often achieved using GNSS sensors, such as receivers with RTK corrections (Thuilot et al., 2002). However, despite the recent improvements of GPS receivers' precision, harsh environmental conditions such as large canopies may decrease the reliability of GNSS sensors inside rows with thick vegetation (Kabir et al., 2016) especially during spring and summer, making GPS-free approaches competitive.
A valid alternative resides in Deep Learning (DL) solutions. In the context of precision agriculture, they are used for many tasks, like fruit detection and counting (Mazzia et al., 2020), land crop classification (Martini et al., 2021), and many others (Ren et al.,

2020). Various DL techniques have been proposed to solve the autonomous navigation problem by overcoming the limits of localization in row crop scenarios, generally combining waypoints generation (Salvetti et al., 2022) with methods of plant segmentation (Aghi et al., 2021) for intra-row control. Other approaches aim at developing navigation policy training Deep Reinforcement Learning agents (Zhu & Zhang, 2021) to guide the robot along the rows, directly mapping images to velocity commands (Martini et al., 2022a).

So far, this work aimed at demonstrating the effectiveness of new tools for artificial data generation in simulation to enhance the development of robust DL-based autonomous navigation algorithms for Unmanned Ground Vehicles (UGVs) in row-based crops.

**Materials and methods**

This section focuses on describing the main contributions of this work. Firstly, the necessary steps to generate the RGB and label mask images of zucchini, lettuce, chard, pear trees, and other fields with Blender (Blender), a free and open-source 3D computer graphics software tool set, are illustrated. Then, the tool to procedurally generate fields according to geometrical parameters is presented and the simulation scenarios obtained in Gazebo (Koenig and Howard, 2004), an open-source 3D robotics simulator, are used to test navigation algorithms.

Dataset generation for semantic segmentation
The first step to generate a realistic synthetic dataset for segmentation is a detailed plant model. The 3D plant models have been developed in Blender using real plant textures and standard dimensions as references. The height of the crop is considered a fundamental factor since it determines the level of obstruction of the GNSS signal, as well as the control strategy and the machine size to adopt. To consider the widest range of crops, three main categories have been identified: low crops, such as lettuce and chard, which have a height of 20-25 cm; medium crops, like zucchini, that reach 60 cm; and tall crops, from vineyard to fruit trees, which can reach $15\text{-}25\cdot 10^{-1}$m and may cause GNSS signal obstruction. Some examples of 3D plant models are shown in Figure 1. A significant factor for successfully testing navigation algorithms in simulated fields is the model of the terrain. Ground irregularity causes the rover to drift or to get stuck. Hence it requires special counteractions in developing a trajectory control strategy. For realistic terrain modeling, a plane is subdivided into multiple polygons in Blender. Then, the polygon vertices are randomly moved, respecting real proportions, to get a bumpy, irregular surface.

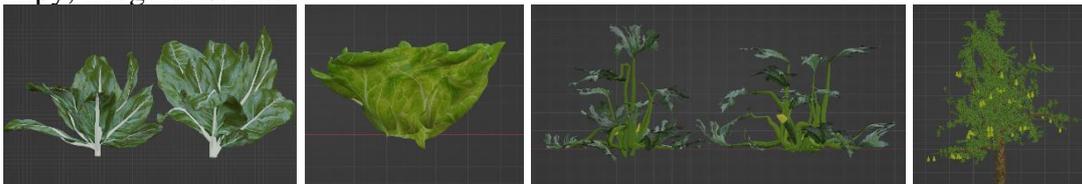

Figure 1: 3D plant models in Blender, from left: chard, lettuce, zucchini, and pear tree.

Moreover, an additional sky model is added to the scene to have realistic visual features for background and illumination. Multiple models are employed to simulate different times of the day, weather, and lighting conditions, to boost the generalization property of the segmentation neural network and avoid unexpected failures.

Once the environmental conditions are set in the field model, a dataset of RGB images and the associate binary segmentation masks can be generated by exploiting the Blender Python scripting functionality, which automatically divides the plants from the rest of the image. In this work, the first version of the dataset is presented, composed of 19456 total image samples for the zucchini field, and 4864 for lettuce, chard, and trees. Nonetheless, each dataset presents four sub-datasets that differ due to the background. Cloudy and sunny skies, diverse lighting, and shadow conditions are considered. For each sub-dataset, the camera pose, both position and orientation, has been changed to acquire diversified image samples along the whole field. The full dataset is available at https://naspic4ser.polito.it/files/sharing/LeqCGJYp6.

Procedural field generation

To enhance the dataset generation and validation process in a simulated environment, a tool to procedurally generate complete fields both as Blender models and Gazebo worlds was developed. The proposed algorithm can generate a generic field with a parametric approach to rapidly test the developed navigation algorithms in various scenarios. Only a terrain and a small set of plant models are required as input, allowing the user to reduce the time spent modeling the entire field drastically.

Several parameters can be used to model the most relevant characteristics of a row crop field. The primary features are the length and the number of rows, which regulate the field dimensions. Crops may also be organized in groups of rows where the inter-row space between two groups is wider than the space between two rows in the same group. The following notation is therefore defined for clarity:

- $d_{rr}$: row to row distance (in a group).
- $d_{RR}$: distance between two groups of rows.
- $d_{pp}$: plant to plant distance in the same row.

Other parameters regulate the plant spacing $d_{pp}$, for instance, the average plant footprint and a scaling factor that can be used to stretch the entire field. The code of the tool is available at the repo https://github.com/PIC4SeR/AutomaticRowCropGenerator.

Simulation environments for autonomous navigation

The complete field models are translated into Gazebo worlds to easily evaluate autonomous navigation algorithms in simulation. Since the scope of the virtual framework is navigation, a simulation with complex visual meshes may result in an unnecessarily high computational cost. Hence, Blender models for plants and terrain are simplified and exported in Object format (.obj). This process has been conducted to generate the meshes for creating Gazebo worlds of zucchini, lettuce, chard, pear trees, and vineyards. Table 1 shows the resulting geometric features of the generated fields. Terrain, plant, and field geometry are considered field's the most relevant descriptive factors. Moreover, other important realistic features have been embedded in the fields, such as small obstacles like fallen branches or stones and terrain slope.

Table 1. Geometric features of the row-based fields realized for zucchini, lettuce, chard, and pear trees. ΔH indicates the maximum irregularity in the terrain. All the values are expressed in meters except for the number of rows.

| | Terrain | | | Plant | | | Field | | | |
|---|---|---|---|---|---|---|---|---|---|---|
| | Length | Width | $|\Delta H|$ | Length | Width | Height | $d_{rr}$ | $d_{RR}$ | $d_{pp}$ | Rows |
| Zucchini | 60 | 38 | 0.2 | 0.82 | 0.9 | 0.6 | 1.8 | 3.6 | 0.7 | 7 |
| Lettuce | 60 | 25 | 0.25 | 0.38 | 0.34 | 0.22 | 0.7 | 1.4 | 0.4 | 3 |
| Chard | 60 | 12 | 0.25 | 0.25 | 0.4 | 0.25 | 0.7 | 1.4 | 0.3 | 3 |
| Pear | 80 | 45 | 0.3 | 1.4 | 2.2 | 3.2 | 5 | 5 | 2.2 | 1 |

**Experiments and Results**

In this section, the quality of the synthetic data generated through simulated fields is validated in two steps. First, the segmentation datasets are used to train and test an efficient deep neural network to segment plant rows in RGB images. Second, the virtual field scenarios in Gazebo are used to test the segmentation-based control algorithm with relevant metrics and set up a new benchmark.

Semantic Segmentation DNN
All the original segmentation datasets have been shuffled over the four sub-datasets and split into a train and a test set. For the largest dataset, the zucchini field, the test set takes 30% of the total dataset, i.e., 5760 images. The test comprises 25% of the dataset for the other crops, i.e., 1216 images, leaving 3648 samples for train and validation.
The architecture of the DNN is the one proposed in the previous work (Aghi et al., 2021): MobileNetV3 (Howard et al., 2019) is used as backbone to extract features from the image, followed by a segmentation head composed of a reduced Atrous Spatial Pyramid Pooling module (Chen et al., 2018). The advantage of this architecture is to exploit rich contextual information from the image at a reduced computational cost. A DNN model is trained on the training dataset obtained for each crop, presenting different backgrounds and camera poses along the entire, using an intersection over unit (IoU) loss:

$$Loss(\theta) = \frac{1}{N}\sum_{i=0}^{N}\left(1 - \frac{\hat{X}_{seg}^i \cap X_{seg}^i}{\hat{X}_{seg}^i \cup X_{seg}^i}\right) \qquad (1)$$

where $\hat{X}_{seg}^i$ is a predicted segmentation mask, and $X_{seg}^i$ is its ground truth mask. Moreover, the backbone has been initialized with pre-trained weights on the CityScapes segmentation datasets (Cordts et al., 2016). Data augmentation transformations such as flip and greyscale have been further applied to images at training time. The best model has been selected according to the best score on the validation set. Quantitative accuracy results on test sets are shown in Table 2, while Figure 2 visually shows the predicted segmentation masks on a collection of test samples. The resulting models are converted in TFLite to enhance inference performance on the CPU (TFLite). All the models present after the conversion a size of 4MB and reach an average inference speed of 60 frame-per-second (fps) on a laptop Intel i7 CPU.

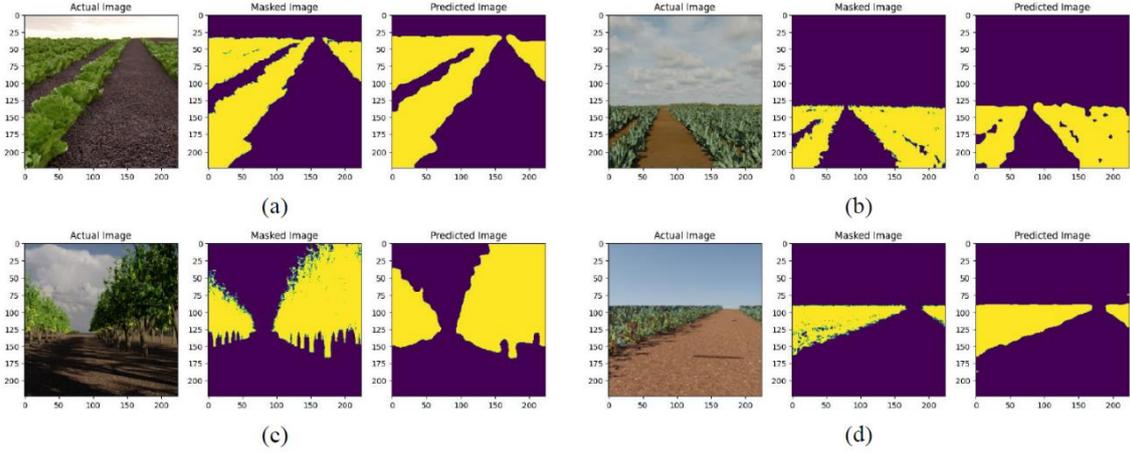

Figure 2. Test samples showing the semantic segmentation results. For each sample, the figure shows the RGB image, the ground truth mask, and the predicted binary mask for: (a) lettuce field with crepuscular light, (b) chard field with cloudy sky, (c) pear trees with sun and long shadows, (d) sunny zucchini field.

Table 2. Semantic segmentation Accuracy and IoU on test sets. *The zucchini dataset includes almost 4 times the number of samples of the others.

|  | Zucchini* | Lettuce | Chard | Pear |
|---|---|---|---|---|
| Accuracy | 0.802 | 0.678 | 0.659 | 0.633 |
| IoU | 197.8 | 93.0 | 6.13 | 145.7 |

Autonomous navigation in row-based crops

The segmentation masks predicted in real-time from the network are used to visually estimate the heading of the rover in the row and compute control commands accordingly. The specific segmentation-based control algorithm is described in (Aghi et al., 2021). The algorithm finds the main cluster of zeros in the segmentation image after summing its values over columns, which corresponds to the free passage in the row. Therefore, it outputs the velocity commands for the rover according to the control laws:

$$\omega_z = -\omega_{z,gain} \cdot d \quad (2) \qquad v_x = v_{x,max} \cdot \left[1 - \left[d^2 / \left(\frac{w}{2}\right)^2\right]\right] \quad (3)$$

where $w$ is the width of the image, $x_c$ is the target cluster's center coordinate, and $d$ is defined as $d = x_c - \frac{w}{2}$. The velocity gains are set equal to $\omega_{z,gain} = 3$ and $v_{x,max} = 1$. All the tests are performed on a 20m long path inside the row, with a maximum linear velocity of 0.5 [m/s] and a maximum angular velocity of 1.0 [rad/s]. The metrics used to evaluate the navigation in the fields are chosen to test the quality of the trajectory of the rover, and the obtained results are gathered in Table 3. The Mean Squared Error (MSE) and the Mean Absolute Error (MAE) are computed between the trajectory followed by the rover and the target one (passing at the center of the row). Moreover, to evaluate the angular oscillation of the rover, the standard deviation of the angular velocity commands and the Cumulative Heading Average (CHA) are recorded, where CHA is defined as $CHA = \frac{1}{N}\sum_{i=0}^{N} \arctan\left(\frac{y_i}{x_i}\right)$. Here, $y_i$ and $x_i$ represents respectively lateral and frontal deviation of a target point with respect to the rover reference frame, for *N* temporal pose samples constituting the navigation travel. For each tested navigation scenario, the goal is a point located centrally at the end of the row. The episodic testing has been performed with the PIC4rl-gym package for automatic metrics calculation (Martini et al., 2022b).

Table 3. Results obtained from navigation tests performed on crops field. Cumulative Heading Average (CHA), Mean Squared Error (MSE), Mean Absolute Error (MAE), and angular velocity Standard Deviation (ω Std Dev) are used as metrics.

|  | CHA [rad] | MAE [m] | MSE [m] | ω Std Dev [rad/s] |
|---|---|---|---|---|
| Zucchini | -0.0346 | 0.1167 | 0.0189 | 0.0428 |
| Lettuce | 0.0474 | 0.1209 | 0.0204 | 0.0224 |
| Chard | -0.0056 | 0.0145 | 0.0004 | 0.0224 |
| Pear | -0.0005 | 0.0058 | 3.8349 | 0.0028 |
| Vineyard | 0.0053 | 0.0330 | 0.0014 | 0.0255 |

The segmentation DNN shows a successful performance in all the scenarios, and the navigation task is completely accomplished. However, the lettuce field presents a higher difficulty level due to the strong irregularity of the terrain combined with stone obstacles, and a narrow space to adjust the trajectory from sudden drifts, as also emerges from the higher errors in the navigation results. Figures 3 and 4 show a Jackal UGV (Clearpath Jackal UGV) inside the Gazebo worlds.

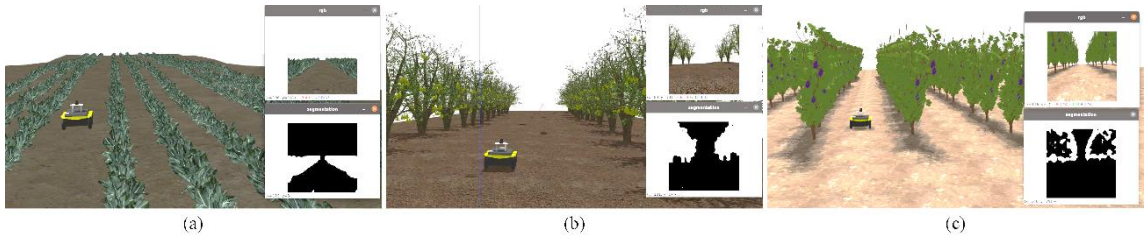

Figure 3. Jackal UGV in chard (a), pear trees (b) and vineyard (c) fields in Gazebo from the robot perspective and current image with the predicted segmentation mask.

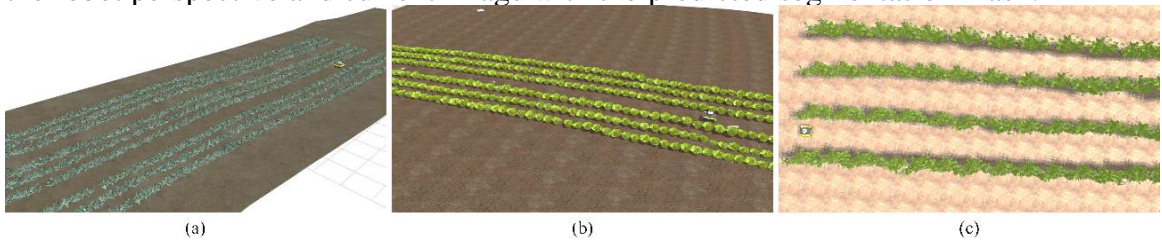

Figure 4. UGV in chard (a), lettuce (b) and vineyard (c) fields in Gazebo from above.

**Conclusion**

This work presents a novel synthetic dataset for semantic segmentation in row-based crops, together with a collection of virtual scenarios for fast development and evaluation of navigation algorithms. A state-of-the-art segmentation-based controller has been developed to validate the dataset and evaluate the simulated scenarios with relevant metrics. Results demonstrate the quality of the generated data and the validity of testing in simulation visual-based navigation algorithms. Future work will extend the dataset to new plants and terrain conditions and will consider the testing of the segmentation models on real crops datasets.

**Acknowledgements**